# Investigating the Role of Explainability and AI Literacy in User Compliance

Niklas Kühl, University of Bayreuth

Christian Meske, Ruhr University Bochum

Max Nitsche, IBM Germany

Jodie Lobana, McMaster University

**Abstract.** AI is becoming increasingly common across different domains. However, as sophisticated AI-based systems are often black-boxed, rendering the decision-making logic opaque, users find it challenging to comply with their recommendations. Although researchers are investigating Explainable AI (XAI) to increase the transparency of the underlying machine learning models, it is unclear what types of explanations are effective and what other factors increase compliance. To better understand the interplay of these factors, we conducted an experiment with 562 participants, who were presented with the recommendations of an AI and two different types of XAI. We find that users' compliance increases with the introduction of XAI but is also affected by AI literacy. We also find that the relationships between AI literacy, XAI, and users' compliance are mediated by the users' mental model of AI. Our study has several implications for successfully designing AI-based systems utilizing XAI.

## 1 INTRODUCTION

With the wide adoption of AI in people's everyday lives, businesses, and society, it has also become ubiquitous and provides increasing automation capabilities in numerous use cases. Such use cases range from credit scoring [21] to medical diagnosis [19]. AI is driven by machine learning (ML) models [27]. They serve as the key enablers—and have become increasingly sophisticated, and, for specific tasks, AI has even been shown to be capable of outperforming humans [18]. A major drawback of these increasingly complex algorithms is that they often appear as "black boxes" since it is hard or impossible for humans to understand the learned concepts and decision-making processes of such algorithms [14]. Explainable AI (XAI) aims to address this limitation by providing human-understandable insights into the inner workings of the underlying algorithms [39]. The reasons to use XAI to improve the limited understanding and transparency of ML models are manifold. With a growing number of applications that use AI and its rising potential, there is inevitably an increase in the number of use cases capable of posing potential risks for humans and society. For instance, unintended automatic decisions in AI-supported driving or autopilots in aviation can have severe consequences [23]. Similarly, AI supporting medical diagnosis can have consequential effects on individuals [41]. Being able to understand and explain the use of ML models is, therefore important in critical AI-based IS [49]. The problem of missing transparency arises particularly with AI that internally uses black-box learning models. One often does not know why the system concluded in a particular manner—for instance, the decision to deny a loan in automated credit scoring. XAI aims to mitigate this aspect. In fact, there are also legal requirements to use fully transparent algorithms, as the European Union passed regulations on automatic decision making and rights to explanation [17]. A factor that is often neglected when discussing all these use cases for XAI is the factor of the individuals who work with the AI. Different techniques of providing insights into the inner workings of an AI's underlying model differ in their complexity, require more or less user expertise or experience, and more or less time to understand. On the other hand, different target audiences might have different needs and requirements for a specific technique. For example, users with more experience in AI, i.e., a higher "AI Literacy" [34], might have different demands and expectations of a technique than users with no experience. The target audience can be characterized by individuals' background and experience. The background and experience of individuals help form their internal representations, referred to as mental models (MMs), which further impact their attitude towards a particular system [22]—in our case, AI.

Our study analyzes the relationship between different XAI types, AI Literacy, MM, and compliance with AI. As recent work stresses, more insights on users' compliance with AI recommendations are needed to understand better the interplay of users' backgrounds and their reliance on AI [4,5,12].We use the notion of MMs to (1) elaborate on how different backgrounds and experiences with AI (AI Literacy) affect the usage of AI and XAI, and (2) how different types of XAI, in turn, influence the MM. Finally, we are interested in studying the effect on the user's compliance with AI when users are presented with different types of XAI techniques. The overall goal of this research is to understand better the relationship between users' AI Literacy, different XAI techniques, and MMs with users' compliance with AI recommendations. We provide first insights that support the tailoring of explanation techniques to users of systems based on their background and experience. Such personalized explanations ("PXAI") can ultimately support and improve users' decision-making processes and their compliance with the AI systems' recommendations.

To research this interplay, we conduct a quantitative study to assess how individuals' AI Literacy, their MMs, together with different types of explanation techniques affect the individuals' compliance with an AI-based information system. As part of this study, we developed our own AI artifact, which can predict the age of people only based on a photograph—and then present participants with the recommendations and their potential explanations of our artifact.

The remainder of this work is structured as follows. First, we set the foundations by elaborating on the terminologies XAI and MMs and relate our work to existing studies. Second, we describe the theoretical foundations and our associated research model. Third, we describe our methodology, consisting of the study design, our AI implementation as well as the data collection. Fourth, we elaborate on our results, which we then discuss. Finally, we conclude our work with a summary, its contributions, limitations as well as suggestions for future research.

## 2 FOUNDATIONS AND RELATED WORK

### 2.1 Explainable AI

While there is no universal or formal definition of XAI, a common understanding is that XAI is—in some way—capable of explaining not only what it does but also why. AI learns, acts, or decides by applying ML algorithms. Consequently, AI that can explain itself requires its underlying ML methods to be explainable.

XAI methods can be classified according to the positioning within the typical ML process among three different criteria: pre-model, in-model, and post-model. The main criterion for XAI is its positioning within the typical ML process [36]: before (pre-model), during (in-model), or after (post-model) building the ML model. Pre-model explanations are carried out before any ML is applied. These explanation methods only take the data, which is later used to build an ML model, as input. The idea is to make this data more understandable and provide insights into how the raw data might impact the decision the AI later makes. Since the AI itself is built later in the ML process, these explanation methods are per definition independent (i.e., model-agnostic) of the ML algorithm. In-model explanation techniques produce explanations as a direct side effect of the applied ML model. Hence, in addition to the trained model, the explanations are an outcome of the model training. ML models that support this explanation are called explainable-by-design or white-box models [39]. Post-model explanations are generated after the model is trained. They can be applied to both white-box models and models which are not intrinsically explainable (black-box models) [8].

## 2.2 Mental Models

First mentioned by Craik [9] in the field of psychology and established by Johnson-Laird [22], the concept of MMs has been picked up by many research fields, including cognitive science, neuroscience, computer science, and IS [55]. MM describes an individual's internal representation of knowledge. This internal representation can be imagined as a small model (of a particular entity or the world in general) that individuals carry in their minds. People construct this internal representation by interacting with their environment. In addition, the MM is influenced by an individual's characteristics, background, experience, and previous knowledge [22]. In the IS context, the model that an individual has in mind about a specific system is of particular interest. Hence, MMs in the IS context can be defined as introduced by McKinney et al. [38] and Vitharana et al. [55]:

Definition 1. A mental model represents a schema of the target system in terms of purpose, functions, processes, and forms.

An individual's MM of a system can change over time as the individual's understanding of the (dynamic) target system changes [20]. The notion of MMs has been employed in various research endeavors in the field of IS. For instance, Santhanam and Sein [48] examine how MMs affect users' proficiency in using the system. In addition, it was used to explore the interaction of individuals with a system, for example, by assessing how individuals' MMs about a system affect the system's usability [13]. The concept of MMs is also leveraged in usability testing [24] and incorporated into the design process of IS [58]. For the remainder of this work, we will be utilizing the definition of MMs provided by Vitharana et al. [55] as cited above.

## 2.3 Related Studies and Positioning

We analyze existing studies with similar objectives and depict their contributions. Lage et al. [31] performed human-grounded experiments to test the effect of explanation lengths and complexities on user comprehension and perceived difficulty. They find that an increased explanation complexity results in longer response times and greater perceived difficulty. Related to this, Bansal et al. [3] investigated how users create a MM of an AI (given basic information such as its error boundary) and tested how this affects the task performance. They show that two AI systems with the exact same accuracy can yield different users' task performance depending on what information and explanations about the AI are given to the user. Similarly, Poursabzi-Sangdeh et al. [42] performed a study to investigate how the number of features and model transparency impact people's decision-making. In their experiment, they asked laypeople to predict the price of apartments or guess the prediction of an ML model. Surprisingly, they observe that "increased transparency hampered people's ability to detect when the model makes a sizable mistake" [42, p.1]. Stumpf et al. [53] highlight the importance of explanations in user interface designs and highlight the positive effect of giving users insight into what the intelligent agent uses as input and output data. They also find that users generally prefer textual instead of visual explanations.

On the other hand, Ribeiro et al. [45] performed a study that demonstrates that visual explanations can help users choose which of two classifiers generalizes better, allow non-expert users to improve the model, and enable users to enable users to identify and describe classifier irregularities. [26] Also, in an application-grounded evaluation in the medical domain, they show significant improvements in experts' ability to interpret and fix an ML model when they are provided with a plot of features that are important for the model. Drawing on the idea of explanations personalized to an individual MM, Kulesza et al. [29] performed an empirical study with 62 participants to explore the effect of MM soundness on personalizing an intelligent agent. The participants were given recommendations of a music recommendation agent, along with varying descriptions of how the agent makes its decisions. They found that participants could quickly understand the recommender system's reasoning and use that understanding to personalize the agent better. In a later study, Kulesza et al.

[30] performed a qualitative study with 17 participants questioning how different ways of explanations impact end users' MMs. The participants were given recommendations of a music recommendation agent (which was based on a decision tree algorithm), along with various types of explanations, including the agent's confidence in each decision and the derivation of a decision based on the decision tree. The authors find that more complete explanations via certain types of information results in a better perception of the benefit of attending to the explanations. They also find that too simple explanations can cause a loss of trust in the recommendation agent. Recent work has looked at explanations and XAI from various angles. As outlined in the previous section, one perspective in related studies is the human factor of XAI. With the insights on the human factor gained from these studies, there have also been endeavors utilizing this knowledge to personalize explanations based on users' preferences [52]. However, the personalizing so far has to be carried out manually by the user by interacting with the system [29,52]. To the best of our knowledge, there is no recent work that considers the users' intrinsic MMs of an AI and XAI and relates it to users' compliance with AI's recommendations. The current study addresses this limitation by examining how different types of explanations influence users' MMs of AI and XAI—and how these MMs affect compliance.

## 3 THEORETICAL FOUNDATIONS AND RESEARCH MODEL

In general, we hypothesize that introducing explanations changes the decision process and, therefore, impacts the results of that decision, more precisely, the compliance with (AI) recommendations. We define compliance according to Fischer et al. 2019 as follows:

**Definition 2.** Compliance is the state of a user being in accordance with a system's recommendation.

Prior research speculated that a) an increase in transparency, as provided by XAI, increases compliance [16] and b) different types of transparency, as available through different types of XAI, influence compliance in different ways [28]. Thus, we hypothesize:

**Hypothesis 1.1:** The introduction of explainability in AI (XAI) changes users' compliance with the recommendations of AI.

**Hypothesis 1.2:** Different types of XAI affect users' compliance differently with the recommendations of AI.

The MMs of individuals shift as they interact with their environments. When individuals are interacting with AI (without XAI), they have one type of mental construct/representation of the AI system. However, as more information is available to them through the explanations provided by XAI, the mental constructs change to accommodate the newly available information. Further, there is evidence that MMs of individuals influence their decision behavior. In the field of IS, it is shown that MMs affect users' compliance with an information system [43]. Hence, we hypothesize:

**Hypothesis 2.1:** The introduction of explainability in AI (XAI) changes users' mental models of AI.

**Hypothesis 2.2:** Users' mental models influence their compliance with the recommendations of AI.

Consistent with the theory of MMs, different people have different attitudes toward a technical artifact, in our case: AI. For example, Wang et al. [57] identified significant effects of computer literacy on fairness perceptions of AI. As opposed to the more general construct of computer literacy, we measure AI Literacy as it applies more directly to our context. We define AI literacy in accordance with Long and Magerko [34].

**Definition 3.** AI literacy is a set of competencies that enables individuals to evaluate AI technologies critically; communicate and collaborate effectively with AI; and use AI as a tool online, at home, and in the workplace

We are interested in whether differences in users' AI literacy, more precisely, their AI skills and the frequency of their AI usage, affect their MMs as well as their compliance with AI's recommendations. Thus, we formulate the following additional hypotheses and summarize our research model in Figure 1:

**Hypothesis 3.1:** Users' AI literacy influences their mental models of AI.
**Hypothesis 3.2:** Users' AI literacy influences their compliance with the recommendations of AI.

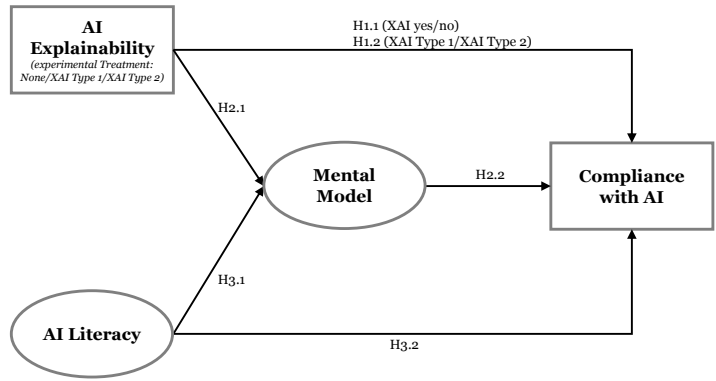

Figure 1: Research Model

## 4 METHODOLOGY

### 4.1 Study Design

To test our hypotheses, we conducted an online experiment, which required participants to perform a simple task both with and without the support of AI and XAI. We showed participants pictures of faces of people and asked them to guess their ages. We chose age estimation because it is a simple task that does not require specific domain knowledge. It is also not too straightforward for humans. It therefore ensures a certain variance in the participants' performance [33]—which AI or XAI could hypothetically influence. Since the work at hand focuses on the general relationship between a participant's MM and compliance, it is essential that the task is not tied to a specific domain. We aim to draw general conclusions about the relationships between MMs and compliance.

For the treatments, we developed and implemented an AI artifact capable of estimating the age of people in pictures. During the first phase, "training", the participants were first provided with no support from the AI for the task. In the second phase of the study, "baseline", our AI artifact supported the participants by simply providing the predicted age estimate. Lastly, in the third phase, "treatment", we introduced our XAI, where participants received one of two distinct types of explanations ("XAI1" or "XAI2").

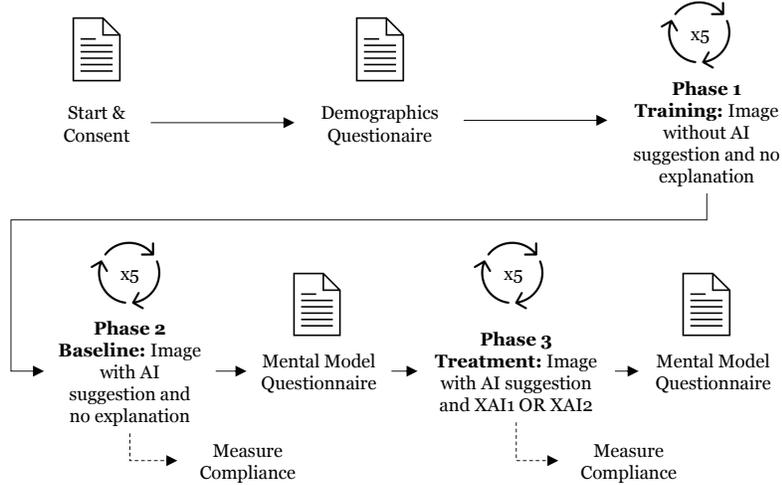

Figure 2: Summary of the experimental process

Therefore, we conduct a study with three distinct groups: a control group, a group for XAI1 ("group XAI1"), and a group for XAI2 ("group XAI2"). The detailed structure of the study goes as follows and is depicted in Figure 2: First, after participants confirm their consent, they are asked initial questions about their background and demographic information, including two items (5-point Likert scale) on rating their AI skills and their AI usage to determine their AI Literacy [34]. In a subsequent step, the participants are introduced to the task of estimating the age of a person on an image. Then, they are asked to guess the age without seeing a prediction of an AI for five consecutive pictures, respectively ("training"). Next, the participants are presented with an introduction to the same task but with the support of an AI ("baseline"). The participants again must guess the age of five people, and, in this part, they are provided with the recommendation of the AI (but not its explanation).

Once they complete this part, they are asked to fill out a questionnaire to determine their MMs based on the subconstructs Goal, Process, and Task as outlined in Vitharana et al. [55]. They are asked to rate their agreement with multiple statements on 5-point Likert scales, where a score of 1 corresponds to "strongly disagree," and a score of 5 denotes "strongly agree". For the control group, the study ends at this point.

For the other two groups with the treatment, the study continues with the same task but with the support of the AI and one type of explanation for each group (either "XAI1" or "XAI2"). Like the previous parts, the participants are introduced to the task, but with an additional introduction to the respective explanation type. The participants again must guess the age of people on five pictures while being provided with the AI recommendation and its explanation. Once this part is complete, the participants fill out the MM questionnaire again. This allows us to track possible changes in the participants' MM after the treatment.

### 4.2 Implementation

The implementation is split into two parts: First, we select a training data set of images of people and their age and use it to develop an AI with an underlying ML model that can predict a person's age based on an input image. In a second step, we implement two types of explanation methods that can provide insights into the decision-making process of the AI.

The decision for a data set for building an AI for age estimation is tightly bound to the current research basis on ML models for age estimation. Age estimation has been of particular interest in the ML community, and many researchers have tackled the task of predicting the age of a person on an image [11,47]. The largest and most popular data set is the IMDB-WIKI data set [47], which we utilize for *training* our AI. For our implementation, we take advantage of the source code published by Serengil [51], with minor adjustments in Python, using the popular *keras* package. The model itself is based on a CNN, which uses the VGG-16 architecture and is pre-trained on the FaceNet database [50]. The network architecture is then adjusted to the age estimation task and our specific data set.

While the IMDB-WIKI data set is widely used as a training basis for age estimation models and the use of existing, published models makes them convenient to use, there are multiple reasons for which the pictures in this data set cannot be used for display (≠ model training) in our study; the quality of the images varies vastly, the ages of the persons are not validated, and the data set contains many pictures of celebrities. Especially the latter could falsify the participants' performance as they might have existing knowledge of the age of a person. Another factor that might have an unintended effect on the study is the fact that the images are taken "in the wild", meaning that there is no standard way of how the people are shown in the image. The people are pictured in various ways, with different poses, facial expressions like smiles or laughter, and clothing like sunglasses, headgear, or jewelry. To address these shortcomings, we use the MORPH data set Feld for model adoption and presentation to the study participants[46]. It has been specifically developed for research purposes and contains the actual age of the people depicted in the pictures. While there are multiple versions of MORPH, the non-commercial release MORPH-II has become a benchmark data set for age recognition [7]. The MORPH-II data set contains unique images of more than 13,000 individuals.

After the model is built, we test its performance in a 10% holdout set, which will also be used within the experiment later. The performance of the models for age prediction is often evaluated by their mean absolute error (MAE). After training and optimization procedures, we reach an MAE of ~2.9 on the MORPH-II data set, which is in line with other researchers [1,54]. This means, on average, our model has an error boundary of +/-3 years when predicting the age.

As stated above, we generate two fundamentally different types of explanations, a chart showing the probability distribution for each age ("XAI1", in-model [2]) and an overlay on an image showing particularly relevant parts of the picture for the AI's prediction ("XAI2", post-model [45]). For the probability distributions, we plot a bar chart that depicts the probabilities—more precisely, the softmax values [56]—for each of the 40 most probable ages. The bars which correspond to the five most probable ages are highlighted in red. An example of such a bar chart, as presented to the participants, is depicted in Figure 3. Note that the probabilities are relatively low, which is rooted in the fact that the probabilities for each of the 101 classes add up to 100%. The model often generates somewhat similar probabilities for ages that are close to each other.

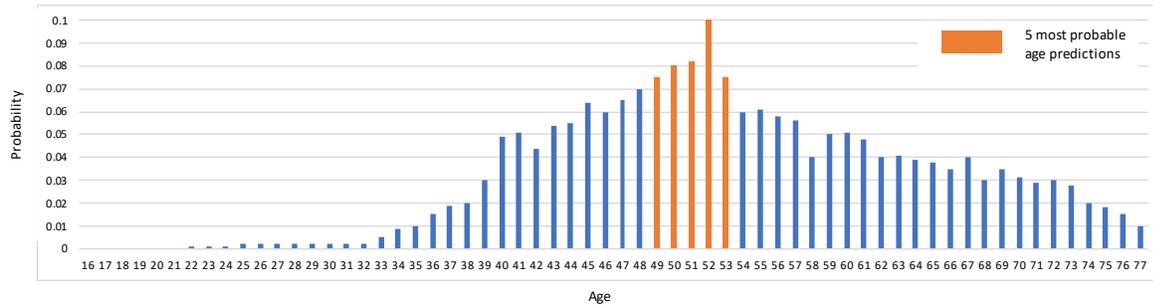

Figure 3: Example of XAI1 Treatment

For "XAI2", the overlay on the images showing relevant parts for the prediction of the CNN, we use the LIME [45] method. It is a post-model method which is applicable to a variety of ML models (and hence our research is not bound to a specific ML model). For each image, we generate overlays showing both the pixels that contribute to the prediction (green overlay) and the ones that worked against the prediction (red overlay). An illustration is provided in Figure 4. The study is implemented with SoSci Survey, a popular online tool to conduct Feld surveys [32].

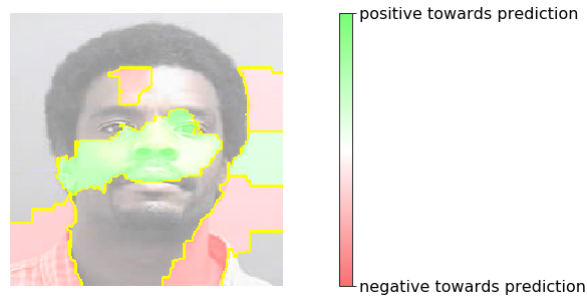

Figure 4: Example of XAI2 Treatment

### 4.3 Data Collection

The study was conducted through the online participant recruitment platform Prolific [40]. We recruited 120 participants for the control group and 221 for group 1 and group 2, respectively. One participant of group 1 did not pass the attention checks and was excluded from the analysis, resulting in 220 participants for group 1. We also made sure that each participant could only participate in one group. The pictures that the participants are presented with are chosen at random from a pool of 200 pictures. In each of the distinct parts with the five images shown to the participants, we add one additional control image of an obviously very young or very old person to be able to later exclude so-called "speeders", as it is a known problem of such online participants recruitment platforms that some participants randomly click through the study [15]. The study took, on average, about 10 minutes for the control group and about 15 minutes for the groups with XAI, respectively. Participants receive an average compensation of 10,35 US dollars per hour. A summary of the participants' characteristics is depicted in Table 1.

Table 1: Summary of Participants' Characteristics

| Participants | Control group | 120 |
|---|---|---|
| | XAI1 | 220 (after exclusion of 1 participant) |
| | XAI2 | 221 |
| Age | | $\mu = 26.0, \sigma = 8.0$ |
| Gender | | 59% male |
| | | 40% female |
| | | 1% prefer not to say / diverse |
| Professional status | | 44% students |
| | | 28% employed full-time |
| | | 9% employed part-time |
| | | 7% self- employed |
| | | 11% unemployed |
| | | 1% other |
| AI Literacy | AI skills | $\mu = 2.5, \sigma = 1.1$ |
| | AI usage | $\mu = 2.5, \sigma = 1.3$ |

## 5 RESULTS

To analyze our data and provide insights into the support of our hypotheses, we conduct different analyses. Our methods range from ANOVA with Post hoc analyses (H1.1) to a full structural equation model (H3.2). We structure this section into three main parts: the effects of XAI on compliance, the effect of XAI on MMs, as well as an analysis of MM as a mediator for compliance.

### 5.1 XAI effects on compliance

We measure compliance as the *negative* mean absolute difference (MAD) between a single user's and the AI's age prediction [35], i.e., as

$$MAD = \frac{\sum |\hat{y}_{AI} - \hat{y}_{user}|}{n}.$$

This means the higher the mean absolute difference (MAD), the lower the compliance. To analyze the effects of XAI on participants' compliance, we regard two scenarios—between-subject and within-subject. First, we compute a between-subject analysis, comparing participants' compliance from different groups. More precisely, we investigate how compliance differed between the control group (CG, no XAI treatment), XAI1, and XAI2. We first compute an ANOVA for insights on general significant differences, revealing a highly significant effect (2.594e-55). This serves as a prerequisite for upcoming Post hoc analyses. We then calculate multiple comparisons of means with Tukey, which gives more insights into the pairwise comparison of the different groups. As shown in Table 2, the change in compliance is significant when we compare the control group with either XAI1 or XAI2. A direct comparison of XAI1 and XAI2 is, however, not significant, and potential differences need to be analyzed further. In summary, the analysis shows that for a between-subject setting, participants' compliance with AI recommendations changes with the introduction of XAI.

Second, we are also interested in a within-subject analysis. Therefore, we compare the differences in compliance for each treatment (XAI1 and XAI2) individually. As the participants were also presented with AI recommendations without XAI for the baseline, we can now observe how the introduction of XAI influenced compliance on an individual level. We conduct two-sided t-tests and compare the compliance before and after the treatment, as shown in Table 3. In both cases, the change of compliance is significant, with p-values of 3.105e-16 for XAI1 and 0.013 for XAI2.

As both between-subject and within-subject analyses show significant results, we can support hypothesis 1.1. From an analysis of the boxplot in Figure 5 on p. 11, we see that compliance not only changes but increases with the introduction of XAI. Thus, our first finding is:

**Finding 1.1:** The introduction of explainability in AI (XAI) *increases* users' compliance with the recommendations of AI.

As our Post Hoc Analysis in Table 2 also reveals, we cannot find significant differences between our treatments regarding XAI1 and XAI2. This means we reject hypothesis 1.2.

Table 2: Significance levels of ANOVA and Multiple Comparison of Means with Tukey for Between-subject perspective

|  |  | Compliance |
|---|---|---|
| ANOVA | All groups compared | *** |
| Multiple Comparison of Means with Tukey | CG ↔ XAI1 | *** |
|  | CG ↔ XAI2 | *** |
|  | XAI1 ↔ XAI2 | n.s. |
| Notes: *p < 0.05, **p < 0.01, ***p < 0.001, n.s. = not significant | | |

Table 3: Two-sided t-test comparing compliance with AI before and after treatment

|  | Compliance |
|---|---|
| AI1 (Baseline, Stage 1) ↔ XAI1 (Stage 2) | *** |
| AI2 (Baseline, Stage 1) ↔ XAI2 (Stage 2) | * |
| Notes: *p < 0.05, **p < 0.01, ***p < 0.001, n.s. = not significant | |

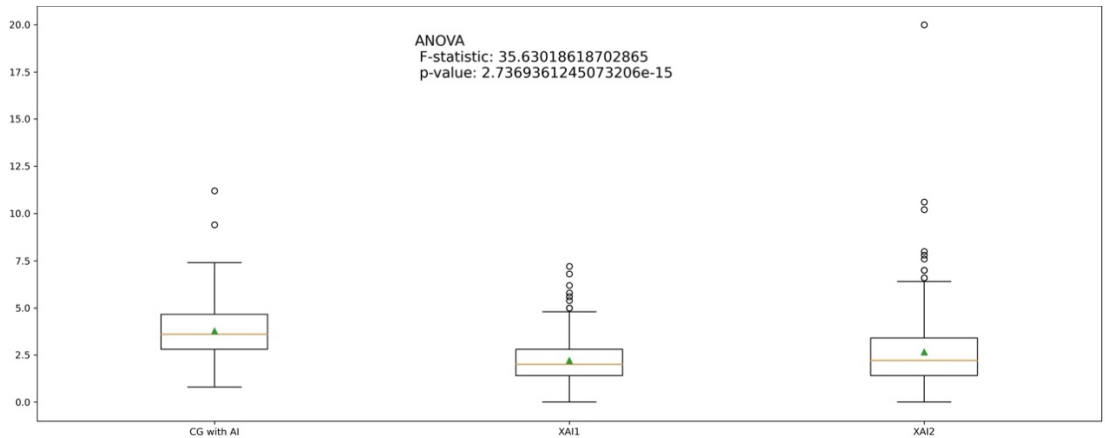
Figure 5: Mean absolute difference (MAD) boxplot of ANOVA for Between-subject perspective of compliance

## 5.2 XAI effects on mental model

We are not only interested in if and how XAI changes participants' compliance with the recommendations of AI, but we also investigate potential changes in their MMs. To do so, we first need to set a few statistical prerequisites to ensure the eligibility of our data. To assess the validity and the reliability of our MM construct, we conduct a confirmatory factor analysis and assess the results with respect to multiple measures. As measures for convergent reliability, we examine Cronbach's alpha (CA), average variance extracted (AVE) and composite reliability (CR).

Table 4: Measurement Information for Latent Factors of Mental Model Construct

|  |  | Control group w/o XAI | | | Treatment with XAI1 | | | Treatment with XAI2 | | |
|---|---|---|---|---|---|---|---|---|---|---|
|  |  | GOAL | TASK | PROC | GOAL | TASK | PROC | GOAL | TASK | PROC |
| 1st order Reliability | CA | 0.825 | 0.950 | 0.894 | 0.813 | 0.945 | 0.904 | 0.876 | 0.939 | 0.902 |
|  | CR | 0.832 | 0.951 | 0.894 | 0.816 | 0.945 | 0.905 | 0.879 | 0.939 | 0.902 |
|  | AVE | 0.625 | 0.866 | 0.739 | 0.597 | 0.851 | 0.762 | 0.709 | 0.838 | 0.755 |
| 2nd order Reliability | CR | MM: 0.705 | | | MM: 0.757 | | | MM: 0.772 | | |
| Notes: CA = Cronbach's alpha, CR = composite reliability, AVE = average variance extracted | | | | | | | | | | |

As depicted in Table 4, for all included cases, the constructs of MM GOAL, TASK and PROCESS, the CA, AVE, and CR are above the recommended thresholds. A confirmatory factor analysis reveals that factor loadings on all items load highly (>0.65) on one factor and with low cross-loadings. These findings demonstrate that our constructs are robust and can be further used in the upcoming analyses. To examine if and how the MMs of participants change with the introduction of XAI with a statistical test, we first need to test for measurement invariance. Measurement invariance is a statistical property of measurement that indicates that the same construct is being measured across some specified groups. Precisely, this means we need to eliminate the possibility that changes in the latent variable between measurement occasions (before and after the treatment) are not attributed to actual change in the latent construct. In the case of an experimental study, this means we need to eliminate a change in the "psychometric" properties of the measurement instrument, i.e., the construct had a different meaning for the participants at measurement occasions. We test the construct of MM, consisting of the subconstructs GOAL, TASK, and PROCESS for metric, scalar, and strict invariance. To compare the means, we require at least scalar invariance [44].

In our case, both metric and scalar invariance are not significant, while strict invariance is significant at the 0.05 level. This means we can compare the latent means of the constructs from before and after the treatment. The results of this comparison are depicted in Table 5.

The values show that the MM changes significantly with the introduction of XAI. For XAI1, the constructs TASK and PROCESS increase by 0.172 and 0.240, respectively. In the case of XAI2, the PROCESS construct changes significantly; more precisely, it increases by 0.3. We can observe no significant change in the GOAL construct, which is, however, not surprising, as the goal of the decision task didn't change. In summary, we can support hypothesis 2.1 and conclude:

Finding 2.1: The introduction of XAI has a positive association with users' mental models of AI.

Table 5: Comparison of latent mean differences across measurement occasions with Wald-Test

| Treatment | Construct | Estimate | SE | z-value | Std.lv | Std.all |
|---|---|---|---|---|---|---|
| XAI1 | GOAL | 0.029 | 0.061 | 0.467 | 0.027 | 0.027 |
| | TASK | 0.168** | 0.055 | 3.036 | 0.172 | 0.172 |
| | PROC | 0.229*** | 0.050 | 4.601 | 0.240 | 0.240 |
| XAI2 | GOAL | 0.063 | 0.072 | 0.873 | 0.061 | 0.061 |
| | TASK | 0.093 | 0.068 | 1.362 | 0.091 | 0.091 |
| | PROC | 0.294*** | 0.058 | 5.051 | 0.300 | 0.300 |

Notes: *p < 0.05, **p < 0.01, ***p < 0.001, SE = standard error, Std.lv = standardized estimates (latent), Std.lv = standardized estimates (all)

### 5.3 Analysis of mental model and AI Literacy

We estimate a full structural equation model (SEM) to better understand the interplay of the different variables considered in our study, we estimate a full structural equation model (SEM). Besides the mental model (MM), compliance (COMP), and the type of XAI (XAI_T), we also include AI Literacy (AILIT). AI Literacy is modeled as the sum of AI Skills and AI Usage. A correlation analysis of the control group reveals that AILIT, MM, and COMP are not considerably correlated (<0.3). Table 6 provides the assessment of our model fit. All indices except for Chi-square are within their required thresholds. The significance of the chi-square test is common for a high amount of observations [25]. The results of the SEM are depicted in Table 7, and the visualization is shown in Figure 6.

Table 6: Model Fit Measures

| CFI | TLI | RMSEA | SRMR | CHISQ |
|---|---|---|---|---|
| 0.981 | 0.975 | 0.054 | 0.035 | 0.00003 |

Notes: CFI = comparative fit index, TLI = Tucker-Lewis index, RMSEA = root mean square error of approximation, SRMR = standardized root mean square residual, CHISQ = Chi-square p-value

Table 7: Results of Model Estimation

| Path | Estimate | SE | z-value | Std.lv | Std.all |
|---|---|---|---|---|---|
| AILIT → MM | 0.096*** | 0.019 | 4.933 | 0.126 | 0.271 |
| AILIT → COMP | -0.156*** | 0.038 | -4.112 | -0.156 | -0.190 |
| (XAI_T → MM | 0.028 | 0.078 | 0-359 | 0.037 | 0.018) |
| XAI_T → COMP | -0.428** | 0.164 | -2.606 | -0.428 | -0.120 |
| MM → COMP | 0.463*** | 0.124 | 3.733 | 0.356 | 0.200 |

Notes: *p < 0.05, **p < 0.01, ***p < 0.001, SE = standard error, Std.lv = standardized estimates (latent), Std.lv = standardized estimates (all)

To investigate our remaining hypotheses, we first examine the effect of the MM on compliance. Here, an increase in the MM led to an increase in compliance (0.2***). Therefore, our results support hypothesis 2.2, and we can conclude:

**Finding 2.2**: Users' mental models influence their compliance with the recommendations of AI.

Next, we assess the relationship between AI Literacy and the MM. Our results show a significant effect between participants' AI Literacy and their MM (0.271***). The higher their AI literacy, i.e., their skills in and usage of AI, the higher their MM. Thus, we can summarize:

**Finding 3.1:** Users' AI literacy influences their mental models of AI.

Finally, we examine the influence of AI Literacy on compliance. As our test shows, AI Literacy has a significant effect on compliance (-0.19***). Interestingly, the relationship is negative, i.e., the higher the AI Literacy of a person, the lower the compliance with the recommendations of the AI. While this might seem counter-intuitive, the theory of *algorithm aversion* [10] supports this result. Research has shown that algorithm aversion can be triggered if people find out algorithms are imperfect [5]. This would make sense in our case as well, as more experienced participants are aware of these imperfections. Thus, we can conclude:

**Finding 3.2:** Users' AI literacy influences their compliance with the recommendations of AI.

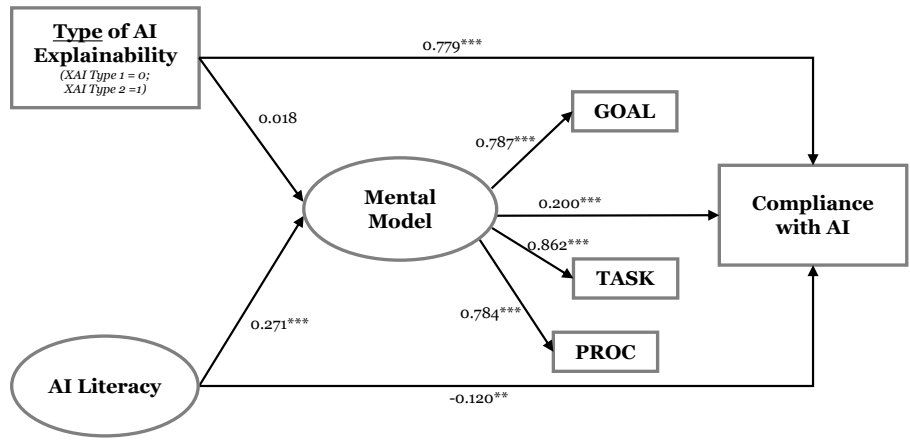

Figure 6: Full Structural Equation Model, Notes: AILIT = AI Literacy, XAI_T = XAI Type, COMP = Compliance, MM = Mental Model; *p < 0.05, **p < 0.01, ***p < 0.001; all estimates are standardized

### 5.4 Subsample analysis for AI literacy

Turning now to the experimental evidence on AI Literacy, we are interested in a deeper analysis of its role. To do so, we created a binary variable by clustering it into two groups for AI Literacy. To end up with balanced group sizes for both items, AI skills and AI usage, we set a cutoff point at "2" (initial Likert scale: "5") and refer to every participant below as "low" AI Literacy (i.e., AI Skills ≤ 2 AND AI Usage ≤ 2) and all other participants above it as "high" (i.e., AI Skills ≥ 3 OR AI Usage ≥ 3). This results in 228 participants with "low" AI Literacy and 333 participants with "high" AI Literacy. Based on this division, we calculate the same SEM as described above but observe how the significances change for the individual cases. The results of the analysis are depicted in Table 8 and interpreted within the upcoming discussion.

Table 8: p-Values for subsample analysis between "low" and "high" AI Literacy participants

| Path | Complete Model | "Low" AI Literacy Participants only | "High" AI Literacy Participants only |
|---|---|---|---|
| AILIT → MM | *** | / | / |
| AILIT → COMP | *** | / | / |
| XAI_T → COMP | *** | n.s. | ** |
| MM → COMP | *** | * | ** |

Notes: *p < 0.05, **p < 0.01, ***p < 0.001, n.s. = not significant, / = not applicable

## 6 DISCUSSION

At a high level, our results show that both AI explanations and the users' AI literacy influence compliance with AI recommendations either directly or indirectly through their impact on users' MMs. A summary of the findings is visualized in Figure 7. The shifts in users' MMs play an essential role in influencing user compliance. When reviewing the results in detail, we find two interesting phenomena. First, we find that AI literacy impacts compliance positively when it is mediated through MM; however, negatively when the impact is measured directly from AI literacy on compliance. Second, we find that the compliance with AI recommendations of users with low AI literacy is not impacted by the explanations provided to them.

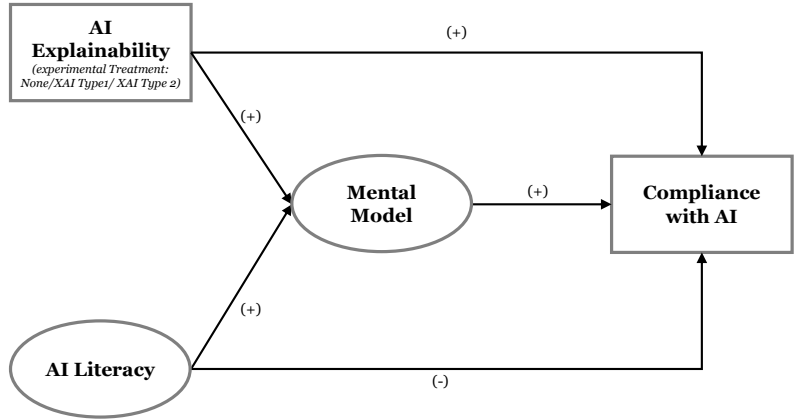

Figure 7: Summary of Findings

The first phenomenon uncovered highlights a paradox where AI Literacy reduces compliance on the one hand but improves MM (which then improves compliance) on the other. This phenomenon is also referred to as inconsistent mediation [37]. We believe that an increase in AI literacy impacts MMs because the increase in skills and experience with AI allows for different, potentially more precise, mental representations of AI (as compared to without that skill and experience). Having said this, in accordance with the algorithm aversion theory, the individuals with a higher level of AI literacy also understand the imperfections in the AI models. This knowledge of imperfections in AI models may decrease their trust in the AI recommendations. With a lower level of trust in AI models, individuals with high AI literacy may tend to trust their own judgment than complying with AI's recommendations. The fact that the SEM model shows an increase in compliance with shifts in MMs but a decrease in compliance directly highlights a tension in the minds of individuals with high AI literacy. They must constantly balance between trusting the precision brought through an AI model or mistrusting the imperfections inherent in the AI model.

The second phenomenon uncovered applies to users with low AI literacy, and the lack of impact of explanations (XA1 and XAI2) on their compliance with AI's recommendations. We made this discovery while conducting subsample analyses of the AI Literacy construct. We found that while the type of XAI (i.e., XAI1 or XAI2) has a significant effect on compliance for participants with high AI Literacy, this does not hold true for participants with low AI Literacy. This means that in terms of compliance, it does not make a difference for participants with low AI Literacy as to which type of XAI is presented to them.

It appears that participants with low AI literacy do not know what to do with the explanations provided to them. Instead, their compliance with AI recommendations is only impacted by their MMs. This finding is a proof point for our call to design more user-centric, more personalized explanations in AI (PXAI). With more personalized explanations, the AI practitioner community can potentially develop explanations that serve users who have no education in AI or statistics to understand even the most basic statistical charts or figures. Since both types of explanations tested within our study used statistical figures and language, it will be useful to conduct future experiments where simple explanations using plain everyday language are utilized to test whether they assist in enhancing compliance of users with low AI literacy.

Beyond the discovery of the above two phenomena, our study further extends the work in IS on MMs. Existing literature in IS has mostly focused on how to measure MMs, while the impact of MMs on actual user behavior is scarce so far. With our findings, we increase the understanding of how MMs influence human-AI-interaction, more specifically, their impact on the compliance of users with AI's recommendations. With this new understanding, we emphasize that users' MM is a variable that researchers and practitioners need to consider when designing and introducing AI.

# 7 CONCLUSION

The importance of AI-based systems is on the rise. However, more exploration into the relationship between humans and AI systems is needed, especially to understand the impact of explanations on users' compliance with the AI recommendations.

In the current study, we elaborate on the relationship between different explainable AI (XAI) methods, users' AI Literacy, MMSs, and compliance with AI recommendations. We layout a research model and an experimental survey setup. We perform a study with 562 participants who estimate the age of multiple persons—once with the help of an AI and once with different treatments of XAI.

Our overall results show that people's compliance with the recommendation of AI increases with the introduction of XAI. Furthermore, we demonstrate that the introduction of XAI changes users' MMs of AI. As analyzed with our full structured equation model, the mental model, in turn, significantly influences users' compliance with the recommendation of AI as well. As MMs originate from the background and experience of people, it is not surprising that their AI Literacy, i.e., their AI skills and usage, influences their compliance with the recommendations of AI as well. In a subsequent analysis, we even find that by differentiating participants into "low" and "high" AI Literacy groups, we can identify that XAI plays different roles for these groups; The type of XAI has no effect on the compliance of participants with low AI Literacy. However, for participants with high AI Literacy, the type of XAI played a significant role.

With these insights, we contribute to the body of knowledge by shedding more light on the relationships between XAI and compliance and the related personal characteristics of the users. We show the importance of personalizing XAI as a function of users' background and experience, i.e., their AI Literacy. We believe this article should start a debate on the necessity of personalized XAI (PXAI). For instance, in the case of medicine, certain types of XAI—like the visual explanation XAI2 from our treatment—might help doctors better understand the recommendation of an AI-based systems. However, when presenting explanations to different patients, different XAI techniques might be required, as their MMs and expertise are probably at different levels. Therefore, we believe PXAI should be the next frontier in user-centric XAI research.

The generalizability of these results is subject to certain limitations. For instance, other relationships might exist that we did not model in our setup. An additional restrictive factor A limitation of this study is the fact that we only included one use case with two different types of XAI. Future work needs to implement additional use cases, especially within specialized domains like medicine, and also study the impact of other XAI techniques. Such work would further deepen our understanding of the influence of XAI on compliance—and might also help to shed more light on the role of the mental model. A promising field of research lies ahead.